\documentclass[journal]{IEEEtran}

\usepackage{amsmath,amsfonts,amssymb,mathtools,amsthm,bbm,multirow,makecell}
\usepackage{algorithm}
\usepackage{algorithmic}
\usepackage{graphicx}
\usepackage{booktabs}
\usepackage{url}
\usepackage{verbatim}
\usepackage{graphicx}

\usepackage[utf8]{inputenc} 
\usepackage[T1]{fontenc}    
\usepackage{hyperref}       
\usepackage{url}            
\usepackage{booktabs}       
\usepackage{amsfonts}       
\usepackage{nicefrac}       
\usepackage{microtype}      
\usepackage{xcolor}         

\usepackage{microtype}
\usepackage{graphicx}
\usepackage{subcaption}
\usepackage{booktabs} 
\usepackage{url}
\usepackage{placeins}

\usepackage{amsmath}
\usepackage{amssymb}
\usepackage{mathtools}
\usepackage{amsthm}
\usepackage{bbm}
\usepackage{multirow}
\usepackage{makecell}

\usepackage{algorithm}
\usepackage{algorithmic}

\usepackage[capitalize,noabbrev]{cleveref}

\theoremstyle{plain}

\theoremstyle{definition}

\theoremstyle{remark}

\begin{document}

\title{Hyperspectral Image Generation with Unmixing Guided Diffusion Model}


\author{Shiyu Shen, Bin Pan, Ziye Zhang and Zhenwei Shi

    \thanks{The work was supported by the National Key Research and Development Program of China under Grant 2022YFA1003800, the National Natural Science Foundation of China under the Grant 62125102, and the Fundamental Research Funds for the Central Universities under grant 63243074. \emph{(Corresponding author: Bin Pan)}}

    \thanks{Shiyu Shen, Bin Pan (corresponding author) and Ziye Zhang are with the School of Statistics and Data Science, KLMDASR, LEBPS, and LPMC, Nankai University, Tianjin 300071, China
        (e-mail: shenshiyu@mail.nankai.edu.cn; panbin@nankai.edu.cn; zhangziye@mail.nankai.edu.cn).}
    \thanks{Zhenwei Shi is with the Image Processing Center,
        School of Astronautics, Beihang University, Beijing 100191, China (e-mail: shitianyang@buaa.edu.cn; shizhenwei@buaa.edu.cn).}
}

\markboth{under review}%
{Shell \MakeLowercase{\textit{et al.}}: A Sample Article Using IEEEtran.cls for IEEE Journals}
\maketitle

\begin{abstract}
        We address hyperspectral image (HSI) synthesis, a problem that has garnered growing interest yet remains constrained by the conditional generative paradigms that limit sample diversity. While diffusion models have emerged as a state-of-the-art solution for high-fidelity image generation, their direct extension from RGB to hyperspectral domains is challenged by the high spectral dimensionality and strict physical constraints inherent to HSIs. To overcome the challenges, we introduce a diffusion framework explicitly guided by hyperspectral unmixing. The approach integrates two collaborative components: (i) an unmixing autoencoder that projects generation from the image domain into a low-dimensional abundance manifold, thereby reducing computational burden while maintaining spectral fidelity; and (ii) an abundance diffusion process that enforces non-negativity and sum-to-one constraints, ensuring physical consistency of the synthesized data. We further propose two evaluation metrics tailored to hyperspectral characteristics. Comprehensive experiments, assessed with both conventional measures and the proposed metrics, demonstrate that our method produces HSIs with both high quality and diversity, advancing the state of the art in hyperspectral data generation.

\end{abstract}


\section{Introduction}
\IEEEPARstart{H}{yperspectral} image (HSI) generation represents a critical need in remote sensing. While HSIs enable a wide range of downstream applications \cite{11077992,10856524}, their acquisition remains fundamentally limited by current sensor capabilities \cite{li2022dual}, resulting in a paucity of large-scale, high-fidelity datasets. This data scarcity continues to challenge methodological progress and robust evaluation in hyperspectral research \cite{10476653}. In light of these limitations, the synthesis of diverse, realistic HSIs is emerging as a pivotal yet comparatively underexplored direction for advancing the field.

However, existing approaches are predominantly conditional and thus unsuitable for scalable data expansion \cite{he2023spectral}. These approaches typically employ conditional constraints, such as RGB images or segmentation maps, to guide the distribution of generated hyperspectral images \cite{zhang2022survey,10856454,10483535}. While effective for spectral super-resolution, these methods fall short of generative modeling, as they prioritize spectral restoration over the comprehensive simulation of the underlying HSI distribution. While such conditioning can improve fidelity, it intrinsically curtails sample diversity and limits the generative capacity required for comprehensive data augmentation.

Unconditional generation, while extensively explored for RGB images \cite{li2023gligen,10496521}, presents unique challenges when applied to HSIs: 
\begin{itemize}
        \item \textbf{High Dimensionality:} HSIs exhibit high spectral dimensionality, with hundreds of bands compared to the three bands in RGB images. This necessitates a fundamentally different model architecture compared to RGB generation models, which typically employ autoencoder structures \cite{kingma2013auto,esser2021taming} to reduce spatial dimensions while enhancing spectral details.
        \item \textbf{Physical Constraints:} HSIs are governed by physical rules \cite{9709645,fu2023hyperspectral}, having limited unique spectral signatures compared to the diversity of RGB images. Rather than attempting to simulate the entire distribution indiscriminately, leveraging these constraints can enhance the fidelity and efficiency of HSI generation.
\end{itemize}

To address these challenges, we integrate hyperspectral unmixing into the generative framework. We project HSIs into the abundance space, which is low-dimensional and governed by the explicit constraints, namely non-negativity and unity \cite{feng2022hyperspectral}. Operating in this physically grounded manifold allows a generator to synthesize HSIs that are both computationally tractable and physically consistent.

To implement the proposed framework, we introduce the Hyperspectral Unmixing Diffusion Model (HUD), which consists of two key components: an Unmixing AutoEncoder (UAE) pair and an Abundance Diffusion Module (ADM). To address the challenge of high dimensionality, we design the UAE module. UAE is initialized using a hyperspectral unmixing algorithm, where the abundances are derived either through rigorously constrained optimization equations or a linear approximation. Reconstruction is performed directly using the endmembers, ensuring efficient dimensionality reduction while preserving essential spectral information. To address the challenge of physical constraint, we construct the ADM module. The diffusion model iteratively generates samples on the abundance space. However, traditional diffusion models rely on Gaussian distributions to describe latent features, which inherently violate the unity and non-negativity constraints of the abundance space. To address this limitation, we shift the abundance space into an unconstrained domain, apply Gaussian-based sampling, and then map the generated samples back to the abundance space. This approach enables the generation of physically consistent and high-fidelity hyperspectral data while maintaining computational efficiency. Moreover, we introduce two novel metrics tailored for HSI generation: point fidelity and block diversity. Experimental results indicate that HUD consistently outperforms existing models on both proposed metrics and traditional metrics.

Our contribution is summarized as the following:
\begin{itemize}
        \item We develop a hyperspectral unmixing guided diffusion model for HSI generation.
        \item We propose an unmixing autoencoder module that shifts the generative task from the image space to the abundance space, reducing dimensionality while preserving fidelity.
        \item We propose an abundance diffusion module, which generates high-quality hyperspectral images that satisfy physical constraints.
\end{itemize}

The rest of this paper is organized as follows. In Section \ref{related_work}, we review related works in the field. Section \ref{method} provides a detailed description of the proposed model. In Section \ref{experiments}, we present the results of real-world experiments. Finally, we conclude the paper in Section \ref{conclusion}.

\section{Related Work\label{related_work}}

We briefly review three relevant areas: diffusion-based generative modeling, HSI generation, and hyperspectral unmixing.

\subsection{Diffusion Models}

Denoising Diffusion Probabilistic Models (DDPMs) \cite{ho2020denoising} have recently become central to state-of-the-art generative modeling. DDPMs define a forward noising process that progressively transforms data into standard Gaussian noise, and a learned reverse-time Markov process that maps noise back to data. The reverse transitions are parameterized by a denoising network, commonly a U-Net \cite{ronneberger2015u}, trained to approximate the score or noise residual.

Latent Diffusion Models (LDMs) \cite{rombach2022high}, popularized by Stable Diffusion, address the computational burden and limited resolution of pixel-space diffusion by operating in a compressed latent space. An autoencoder, typically pretrained on large-scale RGB datasets, projects images to a low-dimensional manifold where diffusion is performed, significantly improving efficiency and achievable resolution relative to pixel-space DDPMs.

\subsection{HSI Generation}
Existing methods for HSI generation typically rely on conditional images \cite{liu2021hyperspectral,10812768}. Depending on the type of conditional input, HSI generation can be categorized into two main approaches: spectral super-resolution \cite{wang2022comprehensive,zhang2023r2h,hang2021spectral} which uses RGB or multi-spectral images as conditions, and synthesis based on semantic segmentation maps \cite{wang2023dcn,chen2023spectraldiff}. A key challenge in HSI generation is the high spectral dimensionality, which differentiates hyperspectral images from RGB images. Regardless of the specific approach, most existing methods for HSI generation focus on generating individual pixels rather than full images. In these models, spatial distribution is conditioned on the input images, while the primary task is the generation of the spectral content for each pixel.

Recently, several studies have explored the use of generative models for HSI generation, including GAN, VAE, and Stable Diffusion \cite{hao2023generative,pang2024hsigene,10327767,10591708,liu2021hsigan,9770778,liu2025specdmhyperspectraldatasetsynthesis,PAN2024102419}. However, most of these approaches still rely on conditional images such as RGB or multi-spectral images during the generation process. This reliance limits the diversity of the generated hyperspectral images, as the spatial distribution is largely determined by the conditional input.

\subsection{Hyperspectral Unmixing}

Hyperspectral unmixing assumes that a HSI can be modeled as a combination of endmembers and their corresponding abundances \cite{feng2022hyperspectral,bioucas2012hyperspectral,borsoi2021spectral}. The fundamental assumption is described by the linear mixing model:
\begin{align}
        Y = AX + \varepsilon,
\end{align}
where \( Y \in R^{c \times h \times w} \) represents the HSI of spatial shape $h\times w$ and spectral shape $c$, \( A \in R^{c\times d} \) is the endmember matrix and $d$ is the number of endmembers, \( X \in R^{d \times h \times w}\) is the abundance matrix, and \( \varepsilon \) is the residual noise. The goal of hyperspectral unmixing is to recover the matrices \( A \) and \( X \) based on \( Y \),  subject to physical constraints.

Recent advancements in hyperspectral unmixing have expanded beyond classical linear models to address complex spectral interactions and nonlinear mixing. Early geometric methods, such as Vertex Component Analysis (VCA) \cite{nascimento2005vertex}, identify endmembers by exploiting spectral data geometry. Statistical approaches like Nonnegative Matrix Factorization \cite{feng2022hyperspectral,9419671,iftene2023partial} enforce physical constraints during decomposition. Sparse unmixing frameworks \cite{shen2023efficient, rasti2021suncnn} leverage sparsity assumptions to improve interpretability. More recently, deep learning-based methods \cite{xu2022deep,gao2024reversible,9444141} have gained prominence, using neural networks to model nonlinear mixing and enhance accuracy under challenging conditions. 

\section{Method\label{method}}


\begin{figure*}[htbp]
        \centering
        \includegraphics[width=1\linewidth]{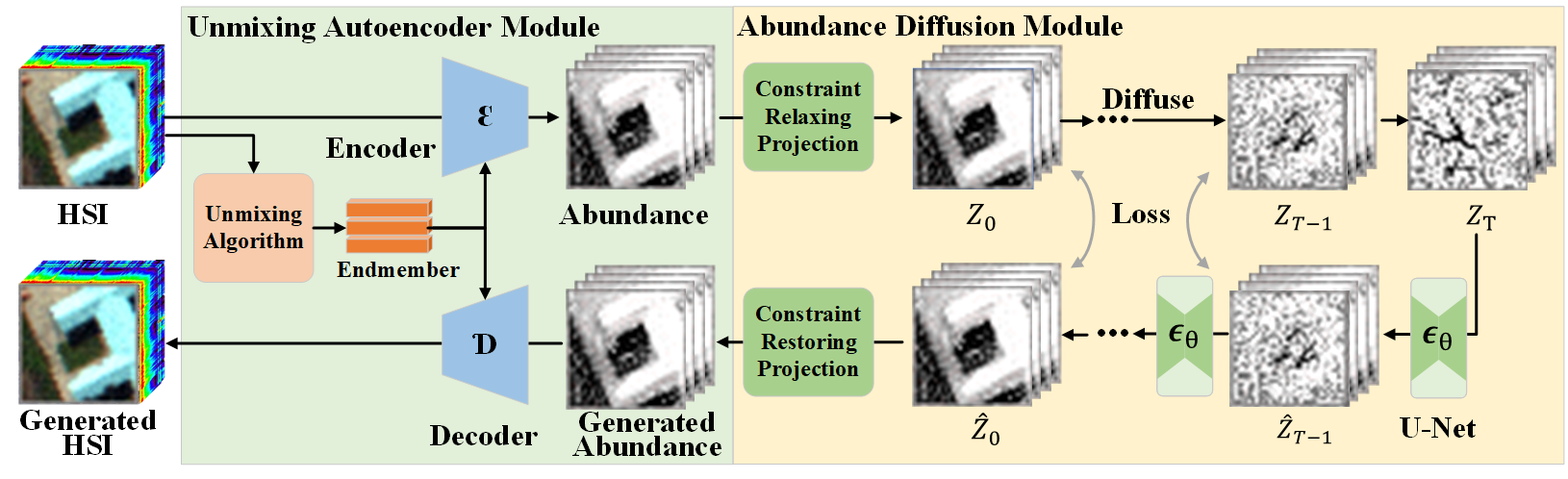}
        \caption{Components of HUD.\label{flow}}
        \label{visualization}
\end{figure*}

In this section, we show the detailed design of the proposed model, including the unmixing autoencoder module and the abundance diffusion module. We first introduce the overall framework of the model, then describe the unmixing autoencoder module and the abundance diffusion module in detail. Finally, we introduce the evaluation metrics for HSI generation.

\subsection{Overall Framework}
The proposed HUD consists of two key modules: the unmixing autoencoder module and the abundance diffusion module. The unmixing autoencoder module reduces the dimensionality of hyperspectral images by projecting them into a low-dimensional abundance space, while the abundance diffusion module generates samples corresponding with physical constraints.

Specifically, given a hyperspectral image dataset, we first extract the endmembers by a hyperspectral unmixing algorithm. Then, the endmembers are used to initialize the autoencoder for the diffusion module. To train the diffusion model, we first encode the hyperspectral images into the abundance space using the unmixing encoder, followed by a projection to relax the constraints of non-negativity and unity. The diffusion model is then trained to learn the distribution of the projected abundance maps. During sampling, we first sample from a standard Gaussian distribution, then apply the denoising process to generate projected abundance maps, and convert them back to the constrained abundance space. Finally, the decoder reconstructs the hyperspectral images from the abundance maps. The overall framework is illustrated in \cref{flow}.

\subsection{Unmixing Autoencoder Module}

The unmixing autoencoder can use any algorithm to extract endmember, and in this paper we use VCA for example. Abundances are subject to two key constraints: non-negativity and unity. Specifically, the abundance values for each pixel must be non-negative and sum to one, reflecting the physical interpretation that each pixel is composed of a mixture of endmembers in varying proportions. Therefore, given \( Y \) and \( A \), \( X\) is usually solved through a constrained optimization problem of the following form:
\begin{align}
        \min ||Y - AX||^2\text{\, s.t.\, }X>0 \text{\, and\, } \sum_i{X_{(i,:,:)}=\mathbbm{1}_{h \times w}}
\end{align}

In addition to the rigorous solution, a linear approximation can also be employed to solve for \( X \). During the extraction of endmembers, many unmixing algorithms inherently incorporate constraints on abundance. As a result, even if these constraints are temporarily relaxed to solve for \( X \), the approximate solution remains close to the rigorously constrained one. The linear approximation to this optimization problem can be expressed as:
\begin{align}
        X = (A^T A)^{-1} A^T Y.
\end{align}
This approximation provides a computationally efficient alternative while maintaining reasonable accuracy, making it suitable for scenarios where computational efficiency is prioritized.

To reconstruct the HSI, we employ a linear projection \(\hat{Y} = AX\), where \(\hat{Y}\) represents the reconstructed HSI. In the proposed model, we initialize a linear layer as the decoder using the endmember matrix \(A\), while the encoder is initialized as \((A^T A)^{-1} A^T\). This initialization can be based on either the linear approximation or the rigorous solution, depending on the desired balance between computational efficiency and accuracy. Although the autoencoder can be further optimized through data-driven training, the limited volume of HSI datasets increases the risk of severe overfitting. Therefore, we recommend using a frozen autoencoder in most scenarios, with fine-tuning reserved for cases where sufficient data is available to ensure robust training without compromising generalization. This approach balances computational efficiency, physical consistency, and the practical constraints of HSI data availability.

\subsection{Abundance Diffusion Module}

We establish a diffusion module on the abundance space. The diffusion module approximates a Markov chain from a Gaussian distribution to the abundance distribution using a U-Net model. It consists of two processes: the diffusion process from the abundance distribution to the Gaussian distribution and the denoising process from the Gaussian distribution to the abundance distribution.

Before training the diffusion model, we introduce a pair of differentiable projections to relax the constraints of non-negativity and unity imposed on the abundance maps. The projections are defined as follows:
\begin{align}
Z[i,j,k] = \ln \left( X[i,j,k] + e^{-\ln(d)-8} \right),
\end{align}
\begin{align}
\hat{X}[i,j,k] = \frac{e^{Z[i,j,k]}}{\sum_{i=1}^{d} e^{Z[i,j,k]}},
\end{align}
where \(Z\) represents the projected feature in an unconstrained space, and \(\hat{X}\) is the reconstructed abundance map that satisfies the original constraints. This mapping is near-lossless for typical magnitudes:
\begin{align}
\hat{X}[i,j,k] = \frac{X[i,j,k] + e^{-\ln(d)-8}}{1 + e^{-8}}.
\end{align}

By operating on \(Z\) instead of \(X\), the constraints are effectively relaxed during the diffusion process, while the final reconstruction \(\hat{X}\) still satisfies the non-negativity and unity conditions. The term \(e^{-\ln(d)-8}\) is added to avoid numerical instability caused by \(\ln(0)\). Since the projection pair is fully differentiable, it can be seamlessly integrated into the end-to-end training of the diffusion model, enabling efficient optimization while preserving the physical consistency of the generated hyperspectral data.

Assume that the distribution of the projected abundance is $Z_0\sim q(Z_0)$, the diffusion process is fixed to a Markov chain that gradually adds Gaussian noise to the $Z_0$ according to a variance schedule $\beta_1,\beta_2,...,\beta_T$:
\begin{align}
         & q(Z_{1:T}|Z_0):= \prod_{t=1}^T{q(Z_t|Z_{t-1})},                                \\
         & q(Z_t|Z_{t-1}):=\mathcal{N} (Z_t;\sqrt{1-\beta_t}Z_{t-1},\beta_t \mathbf{I}  )
\end{align}

The denoising process is the reverse of the diffusion process. The transition is a Gaussian distribution with learnable parameter $\theta$, which starts from $p(Z_T)=\mathcal{N}(Z_T;\mathbf(0),\mathbf{I})$:
\begin{align}
         & p_\theta(Z_{0:T}):=p(X_T)\prod_{t=1}^T{p_\theta(Z_{t-1}|Z_{t})},                   \\
         & p_\theta(Z_{t-1}|Z_t):=\mathcal{N}(Z_{t-1};\mu_\theta(Z_t,t),\Sigma_\theta(Z_t,t))
\end{align}

To train the learnable distribution, we optimize a variational lower bound on the negative log-likelihood. Specifically, we aim to minimize the following objective:
\begin{multline}
        \mathbf{E}[-\log p_\theta(Z_0)]\leqslant \mathbf{E}_q\left[-\log \frac{p_\theta(Z_{0:T})}{q(Z_{1:T}|Z_0)}\right]\\
        =\mathbf{E}_q\left[-\log p(Z_T)-\sum_{t=1}^T\frac{p_\theta(Z_{t-1}|Z_t)}{q(Z_t|Z_{t-1})}\right]
\end{multline}

By notation $\alpha_t := 1-\beta_t$ and $\bar{\alpha_t}:=\prod_{s=1}^t\alpha_s $, the loss term can be rewritten as:
\begin{multline}\label{equation8}
        \mathbf{E}_q[D_{KL}(q(Z_T|Z_0)||p(Z_T))\\
        +\sum_{t=2}^{T}D_{KL}(q(Z_{t-1}|Z_t,Z_0)||p_\theta(Z_{t-1}|Z_t))\\
        -\log p_\theta(Z_0|Z_1)]
\end{multline}
where
\begin{align*}
q(Z_t|Z_0)&=\mathcal{N}(Z_t;\sqrt{\bar{\alpha_t}}Z_0,(1-\bar{\alpha_t})\mathbf{I}),\\
q(Z_{t-1}|Z_t,Z_0) &= \mathcal{N}(Z_{t-1};\tilde{\mu_t}(Z_t,Z_0),\tilde{\beta_t}\mathbf{I} ),\\
\tilde{\mu_t}(Z_t,Z_0)&=\frac{\sqrt{\bar{\alpha}_{t-1}}\beta_t}{1-\bar{\alpha}_t}Z_0+\frac{\sqrt{\alpha_t}(1-\bar{\alpha}_{t-1})}{1-\bar{\alpha}_t}Z_t,\\
\tilde{\beta}_t&=\frac{1-\bar{\alpha}_{t-1}}{1-\bar{\alpha}_t}\beta_t.
\end{align*}

For simplification, we assume that $\Sigma_\theta(Z_t,t)=\sigma^2_t\mathbf{I}$, and use reparameterizing $Z_t(Z_0,\epsilon) = \sqrt{\bar{\alpha}_t}Z_0=\sqrt{1-\bar{\alpha}_t}\epsilon$ for $\epsilon \sim \mathcal{N}(0,\mathbf{I})$, the middle term in \cref{equation8} can be written as:
\begin{align}
        \begin{aligned}
                  & D_{KL}(q(Z_{t-1}|Z_t,Z_0)||p_\theta(Z_{t-1}|Z_t))                                                                                                        \\
                = & \mathbb{E}_q\left[\frac{1}{2\sigma^2_t}||\tilde{\mu}_t(Z_t,Z_0)-\mu_\theta(Z_t,t)||^2\right]+C                                                           \\
                = & \mathbb{E}_{Z_0,\epsilon}\left[\right.                                                                                                                   \\
                  & \frac{1}{2\sigma^2_t}|| \frac{1}{\sqrt{\alpha_t}}(Z_t(Z_0,\epsilon)-\frac{\beta_t}{\sqrt{1-\bar{\alpha}_t}}\epsilon)-\mu_\theta(Z_t(Z_0,\epsilon),t)||^2 \\
                  & \left.\right]
        \end{aligned}
\end{align}

As a result, $\mu_\theta$ should approximate $\frac{1}{\sqrt{\alpha_t}}(Z_t-\frac{\beta_t}{\sqrt{1-\bar{\alpha}_t}}\epsilon)$, so we choose the form of $\mu_\theta$ as:
\begin{align}
        \mu_\theta(Z_t,t)=\frac{1}{\sqrt{\alpha_t}}(Z_t-\frac{\beta_t}{\sqrt{1-\bar{\alpha}_t}}\epsilon_\theta(Z_t,t))
\end{align}
where $\epsilon_\theta$ is represented by a U-Net. Finally, the practical loss function is:
\begin{align}
        L := \mathbf{E}_{t,Z_0,\epsilon}\left[||\epsilon-\epsilon_\theta(\sqrt{\bar{\alpha}_t}Z_0+\sqrt{1-\bar{\alpha}_t}\epsilon,t)||^2\right]
\end{align}
where t is uniformly distributed on $[1:T]$.

After training, we sample from standard Gaussian distribution to get $Z_T$ and follow the denoising process step by step. The training and sampling procedure are shown in \cref{algorithm_training,algorithm_sampling}.

\begin{algorithm}
        \caption{Training\label{algorithm_training}}
        \begin{algorithmic}
                \WHILE{not converged}
                \STATE sample minibatch $Y$
                \STATE $X = \mathcal{E}(Y)$
                \STATE $Z_0 = \ln (X+e^{-\ln (d)-8})$
                \STATE sample $t \sim \mathsf{U}[1:T]$
                \STATE sample $\epsilon \sim \mathcal{N}(0,\mathbf{I})$
                \STATE make backward propagation by $\triangledown_\theta||\epsilon-\epsilon_\theta(\sqrt{\bar{\alpha}_t}Z_0+\sqrt{1-\bar{\alpha}_t}\epsilon,t)||^2$
                \ENDWHILE
        \end{algorithmic}
\end{algorithm}

\begin{algorithm}
        \caption{Sampling\label{algorithm_sampling}}
        \begin{algorithmic}
                \STATE sample $Z_T \sim \mathcal{N}(0,\mathbf{I})$
                \FOR{t in $T,T-1,...,1$}
                \STATE sample $\epsilon \sim \mathcal{N}(0,\mathbf{I})$
                \STATE $Z_{t-1}=\frac{1}{\sqrt{\alpha_t}}(Z_t)-\frac{1-\alpha_t}{\sqrt{1-\bar{\alpha}_t}}\epsilon_\theta(Z_t,t)+\sigma_t\epsilon$
                \ENDFOR
                \STATE $\hat{X} = \text{softmax}(Z_0)$
                \STATE $\hat{Y}=\mathcal{D}(\hat{X})$
                \RETURN $\hat{Y}$
        \end{algorithmic}
\end{algorithm}

\subsection{HSI Specialized Evaluation Metric\label{metrics}}

\begin{figure}[htbp]
        \centering
        \includegraphics[width=1\linewidth]{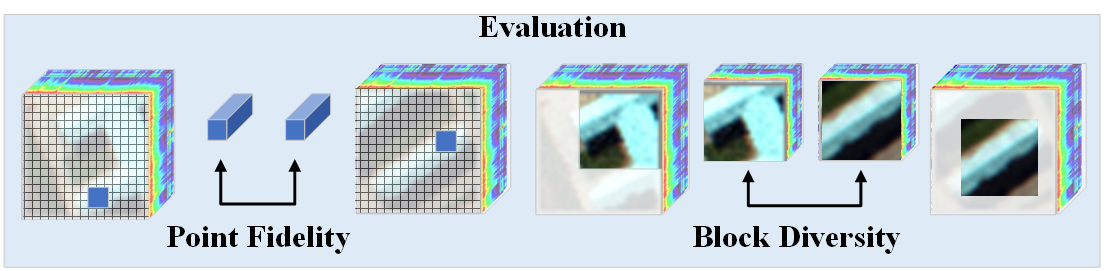}
        \caption{Point fidelity and block diversity\label{evaluation}}
\end{figure}

Existing evaluation metrics for conditional HSI generation primarily focus on measuring the similarity between generated images and ground truth images, which are less effective for assessing the diversity and realism of the unconditionally generated samples. Inception Score (IS) and Fréchet Inception Distance (FID) are common choices in RGB image generation. However, they require a classification model pretrained on large-scale datasets, which is not feasible for HSI due to the limited data size. Moreover, IS and FID are not designed to capture the unique characteristics of HSI data, such as the high spectral dimensionality and limited spatial resolution.

Consequently, we propose two evaluation metrics tailored for HSI: point fidelity $F_p$ and block diversity $D_b$. Point fidelity considers every pixel in the generated image and measures the cosine similarity to the most similar pixel in the original image. To be specific:
\begin{align}
        F_p = \frac{1}{|\hat{Y}|}\sum_{\hat{y}\in \hat{Y}}\max_{y\in Y}\left[\frac{\hat{y}}{||\hat{y}||}*\frac{y}{||y||}\right]
\end{align}
where $Y$ is the real HSI, $\hat{Y}$ is the generated HSI, $y$ and $\hat{y}$ are pixels, and $|\hat{Y}|$ is the amount of pixels. In contrast, block diversity reflects whether the overall distribution of the generated HSI is a direct copy of the original HSI, or if there are differences in the distribution during the generation process. To be specific:
\begin{align}
        D_b = \frac{1}{N_b}\sum_{\hat{y}_b \subset \hat{Y}}\frac{1}{|\hat{y}_b|}\max_{y_b\in X}\left[\frac{\hat{y}_b}{||\hat{y}_b||}*\frac{y_b}{||y_b||}\right]
\end{align}
where $N_b$ is the number of blocks, $y_b$ is a block from the real HSI, $\hat{y_b}$ is a block from the generated HSI. We assume that the generated image size is smaller than the original image because generating a full-sized hyperspectral image in the entire spatial domain is currently infeasible, even with reduced computational demand from the autoencoder.

\section{Experiments\label{experiments}}

In this section, we empirically demonstrate the superiority of HUD. We first introduce the datasets for the experiments, the comparison models, and the selection of hyperparameters. Then, we list images generated by different models. Finally, we analyze the performance of different models using quantitative metrics.

\subsection{Experiments Setup}

We use Indianpines, KSC, Pavia, PaviaU, and Salinas to showcase the generation results. (1) \emph{Indianpines} contains an image scene with spatial size $145\times145$ and 220 bands, covering a wavelength range of $[104-108], [150-163], 220$ nm, having 16 classes. The Band 46,17,11 from the data are chosen for pseudo-color visualization. (2) \emph{KSC} contains an image scene with spatial size $512\times614$ and 176 bands, covering a wavelength range of 400 - 2500 nm, having 13 classes. The Band 28,9,10 from the data are chosen for pseudo-color visualization. (3) \emph{Pavia} contains an image scene with spatial size $1093\times715$ and 102 bands, having 9 classes. The Band 46,27,10 from the data are chosen for pseudo-color visualization. (4) \emph{PaviaU} contains an image scene with spatial size $610\times340$ and 103 bands, having 9 classes. The Band 46,27,10 from the data are chosen for pseudo-color visualization. (5) \emph{Salinas} contains an image scene with spatial size $512\times217$ and 204 bands, covering a wavelength range of $[104-108], [150-163], 220$ nm, having 16 classes. The Band 36,17,11 from the data are chosen for pseudo-color visualization. 

HUD will be compared with the following models: VAE, GAN \cite{gulrajani2017improved}, MPRNet \cite{Zamir_2021_CVPR}, UD \cite{10591708} and UBF \cite{Yu_2024_CVPR}. As there are few hyperspectral image generation algorithms, we adapted the classic models VAE and GAN from RGB generation tasks. Their first convolutional layer, usually a 1x1 conv layer, is resized according to the dimensions of the dataset. The other structures remain unchanged. We also evaluate the generation performance of MPRNet, a state-of-the-art hyperspectral image generation model, but it requires RGB images as conditions. Following the original paper, we directly input the pseudo RGB images to MPRNet. We also compare with UD and UBF, which are the state-of-the-art diffusion based generation models for hyperspectral images. The implementation of these models is based on the official code. We \emph{only} change the dataloaders according to the datasets, and the structures and hyperparameters remain unchanged.

The UAE module is initialized separately for each dataset, and the dimensionality of \(Z\) is determined by the number of categories in the dataset. The UAE is pretrained and does not participate in the training of the diffusion model. The time steps of the diffusion model are set to 1000. Except for the downsampling layer, the structures and hyperparameters of VAE, GAN, and MPRNet are set according to the original papers. All models are trained on a 4090 GPU. We randomly crop 32x32 overlapping sub-images from the original image as training samples.

\subsection{Qualitative Experiments}

In this section, we will showcase the quality of generated images through visualization, including pseudo-color visualization and spectral curve visualization.

\subsubsection{Results on Pseudo-Color Images}
\begin{figure*}[htbp]
        \centering
        \includegraphics[width=0.9\linewidth]{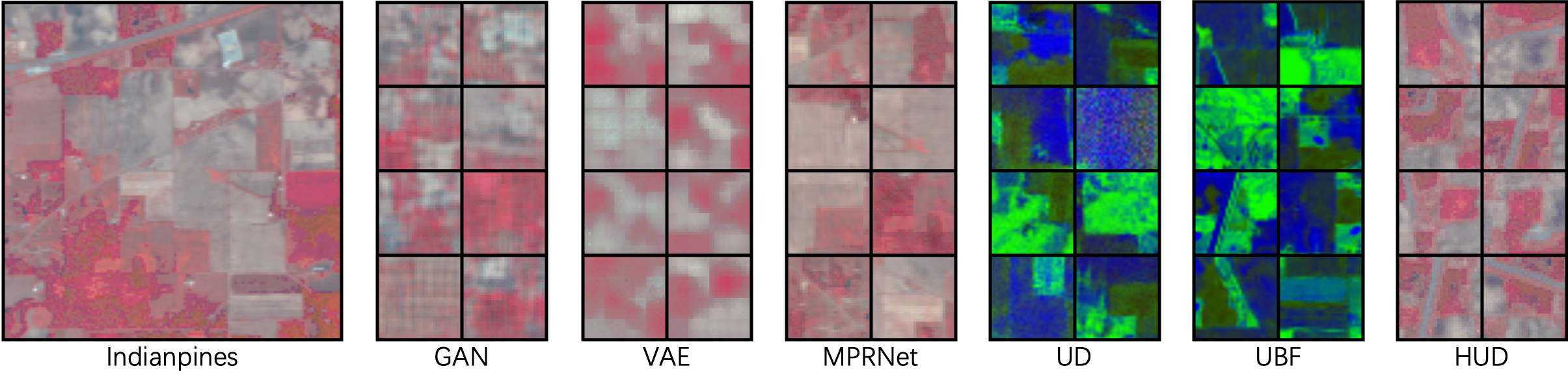}
        \includegraphics[width=0.9\linewidth]{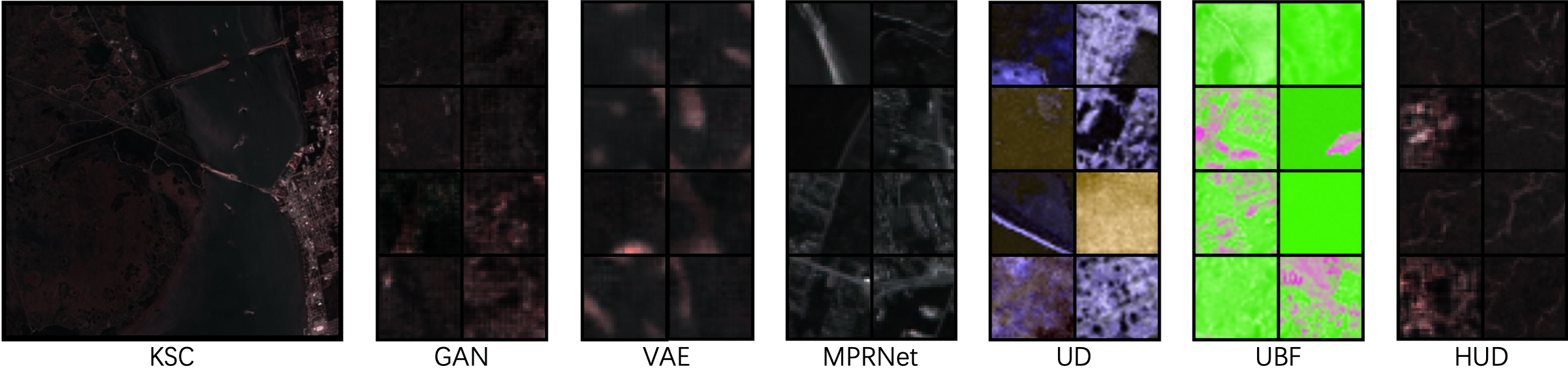}
        \includegraphics[width=0.9\linewidth]{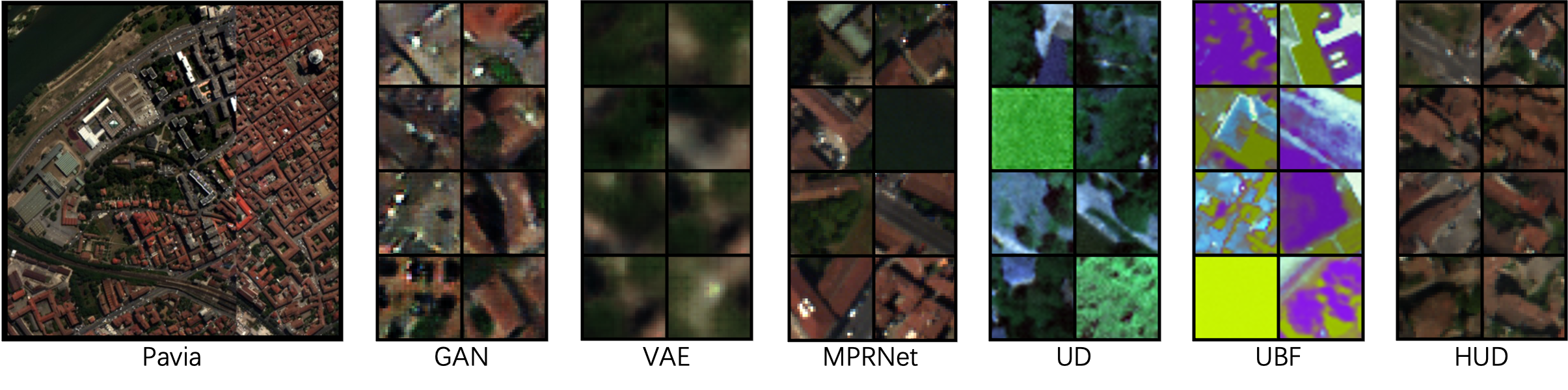}
        \includegraphics[width=0.9\linewidth]{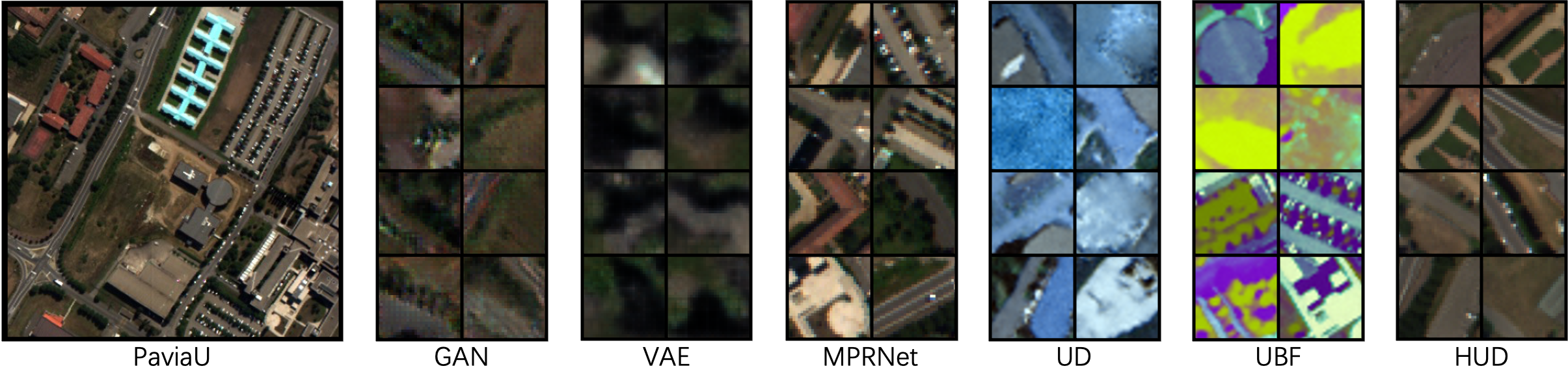}
        \includegraphics[width=0.9\linewidth]{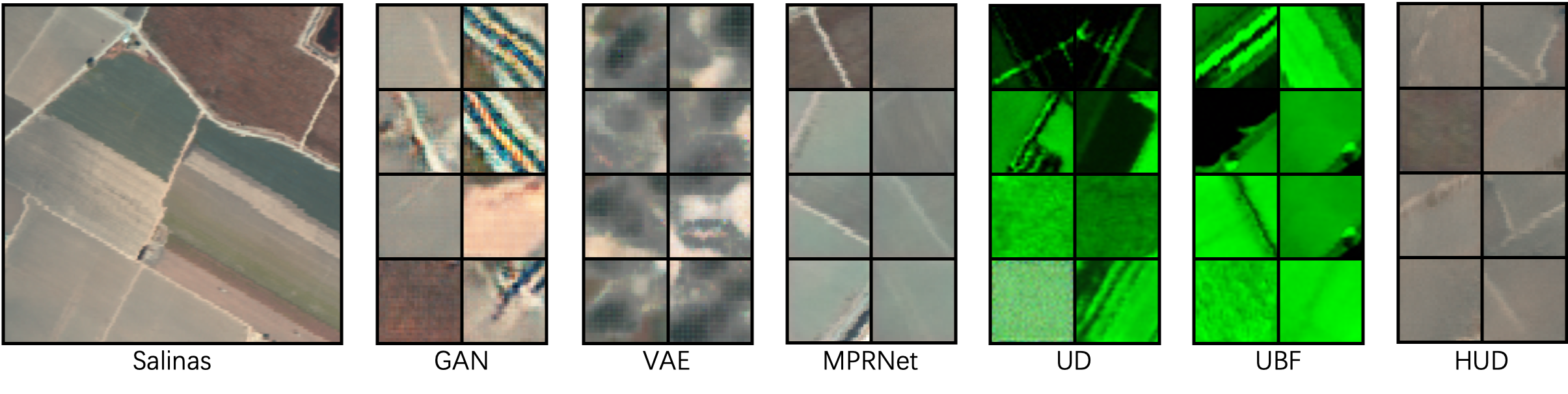}
        \caption{Pseudo-color visualization of real HSI and generated HSI.\label{pseudo_color}}
\end{figure*}
\begin{figure*}[htbp]
        \centering 
        \subfloat[0.13\linewidth][Real\label{Real}]{
                \begin{minipage}[c]{0.13\linewidth}
                        \rotatebox{90}{~~~Indianpines}\includegraphics[width=1\linewidth]{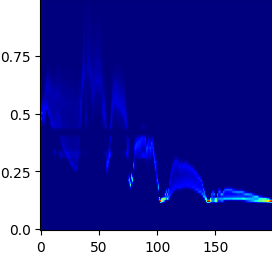}
                        \rotatebox{90}{~~~~~~KSC}\includegraphics[width=1\linewidth]{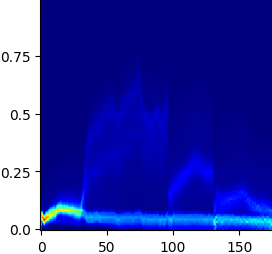}
                        \rotatebox{90}{~~~~Pavia}\includegraphics[width=1\linewidth]{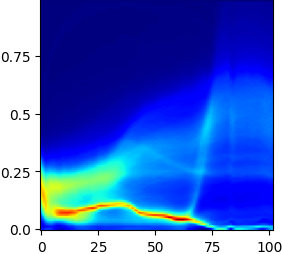}
                        \rotatebox{90}{~~~~PaviaU}\includegraphics[width=1\linewidth]{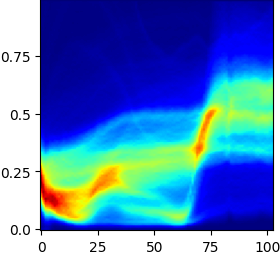}
                        \rotatebox{90}{~~~~Salinas}\includegraphics[width=1\linewidth]{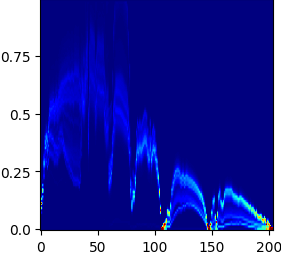}
                \end{minipage}
        }
        \subfloat[0.13\linewidth][GAN\label{GAN}]{
                \begin{minipage}[c]{0.13\linewidth}
                        \includegraphics[width=1\linewidth]{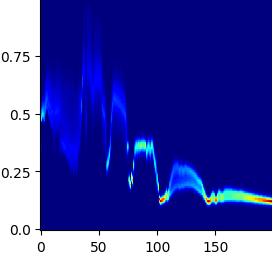}
                        \includegraphics[width=1\linewidth]{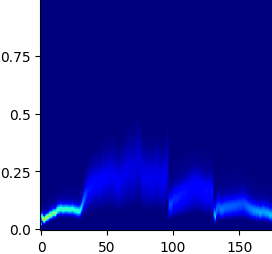}
                        \includegraphics[width=1\linewidth]{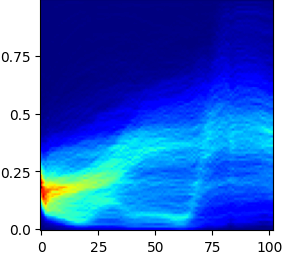}
                        \includegraphics[width=1\linewidth]{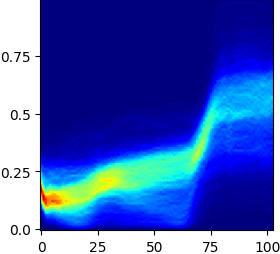}
                        \includegraphics[width=1\linewidth]{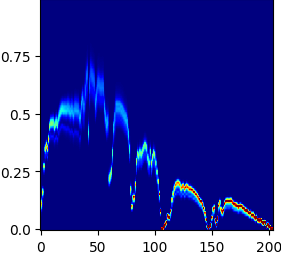}
                \end{minipage}
        }
        \subfloat[0.13\linewidth][VAE\label{VAE}]{
                \begin{minipage}[c]{0.13\linewidth}
                        \includegraphics[width=1\linewidth]{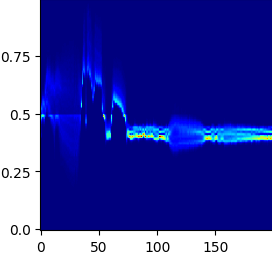}
                        \includegraphics[width=1\linewidth]{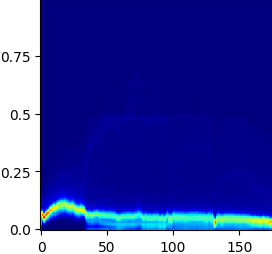}
                        \includegraphics[width=1\linewidth]{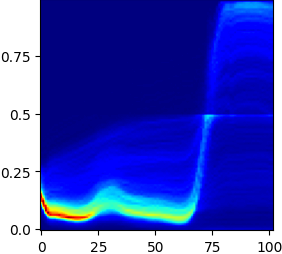}
                        \includegraphics[width=1\linewidth]{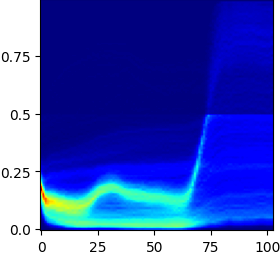}
                        \includegraphics[width=1\linewidth]{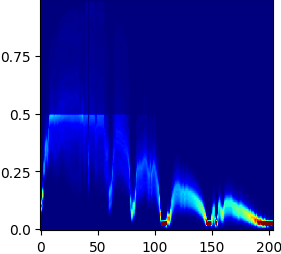}
                \end{minipage}
        }
        \subfloat[0.13\linewidth][MPRNet\label{MPRNet}]{
                \begin{minipage}[c]{0.13\linewidth}
                        \includegraphics[width=1\linewidth]{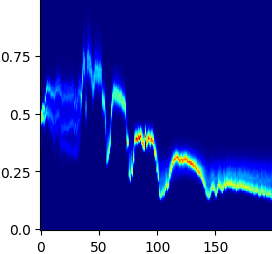}
                        \includegraphics[width=1\linewidth]{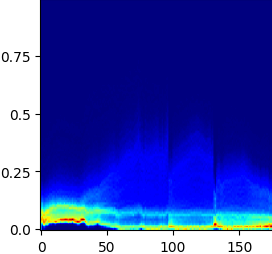}
                        \includegraphics[width=1\linewidth]{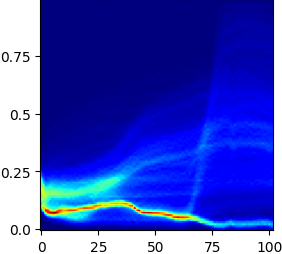}
                        \includegraphics[width=1\linewidth]{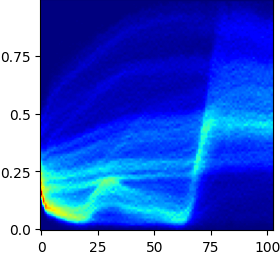}
                        \includegraphics[width=1\linewidth]{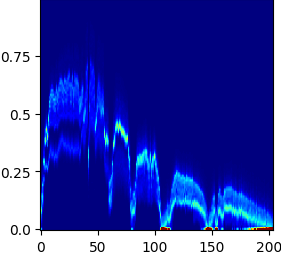}
                \end{minipage}
        }
        \subfloat[0.13\linewidth][UD\label{UD}]{
                \begin{minipage}[c]{0.13\linewidth}
                        \includegraphics[width=1\linewidth]{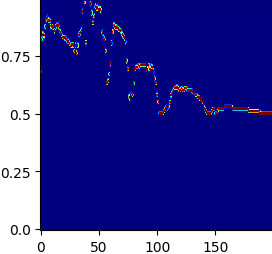}
                        \includegraphics[width=1\linewidth]{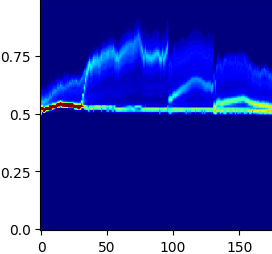}
                        \includegraphics[width=1\linewidth]{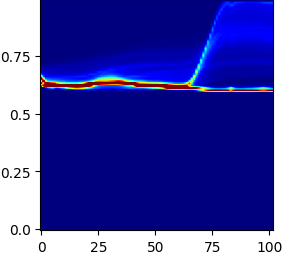}
                        \includegraphics[width=1\linewidth]{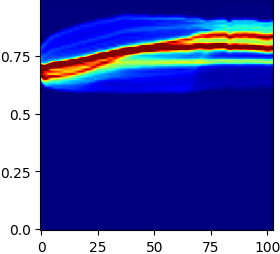}
                        \includegraphics[width=1\linewidth]{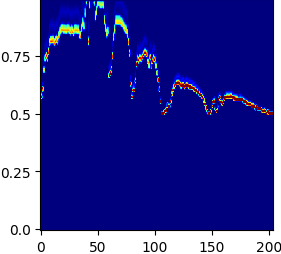}
                \end{minipage}
        }
        \subfloat[0.13\linewidth][UBF\label{UBF}]{
                \begin{minipage}[c]{0.13\linewidth}
                        \includegraphics[width=1\linewidth]{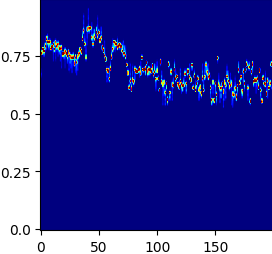}
                        \includegraphics[width=1\linewidth]{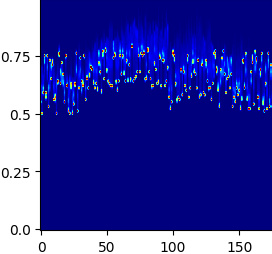}
                        \includegraphics[width=1\linewidth]{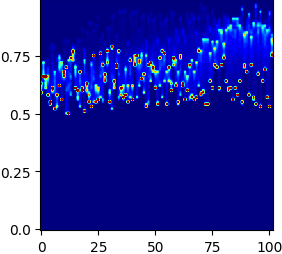}
                        \includegraphics[width=1\linewidth]{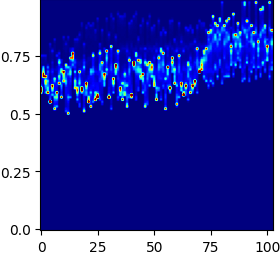}
                        \includegraphics[width=1\linewidth]{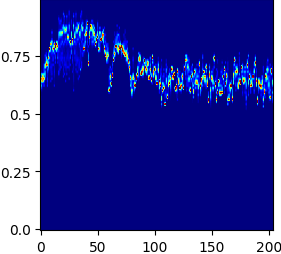}
                \end{minipage}
        }
        \subfloat[0.13\linewidth][HUD\label{HUD}]{
                \begin{minipage}[c]{0.13\linewidth}
                        \includegraphics[width=1\linewidth]{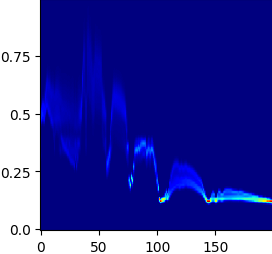}
                        \includegraphics[width=1\linewidth]{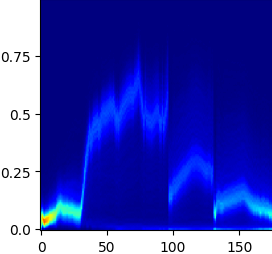}
                        \includegraphics[width=1\linewidth]{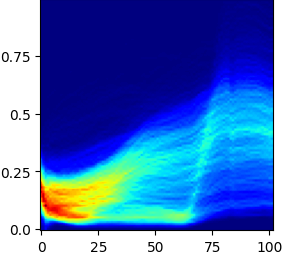}
                        \includegraphics[width=1\linewidth]{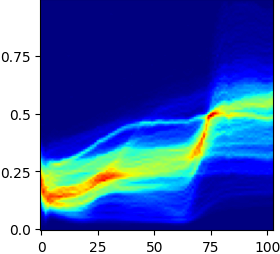}
                        \includegraphics[width=1\linewidth]{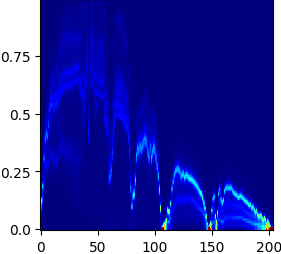}
                \end{minipage}
        }
        
        \caption{Spectral curve distributions of real HSI and generated HSI.\label{spectral_curve}}
\end{figure*}

The pseudo-color visualization of 5 datasets is shown in \cref{pseudo_color}. The first column shows the pseudo-color visualization of the real HSI, and the generated images from different methods are listed on the right side. The pseudo-color visualization can reflect the spatial distribution of HSIs. Specifically, the generated images from GAN have grids overlay on the surface, resembling a mosaic. This problem is probability caused by the downsampling layers in generators and discriminators, which overlook some specific areas because of the pooling or striding operation. VAE directly estimates the distribution of original images, but the scheme limits the performance of the model. Consequently, the generated images from VAE are blurry and lack details. As a spectral super-resolution model, MPRNet generates images with high quality, but the spatial distribution is limited to the original image. Although the autoencoders of UD and UBF are trained with physical constraint normalization, they cannot well restore the spectral information. As a result, UD and UBF have good spatial distributions but poor color visualization. HUD also generates high-quality images, and the spatial distribution is more diverse than MPRNet. For example, there are some new roads and buildings in the generated images of HUD, which are not present in the original images.

\subsubsection{Results on Spectral Curves}

The spectral curve distributions of the real HSI and the generated HSI are shown in \cref{spectral_curve}. We stack all the spectral curves of an HSI to show the spectral distributions. The boundary shows whether the generated spectral curves are consistent with the real HSI, and the density (color) infers the distribution of endmembers. Generally, GAN, MPRNet and HUD generate spectral curves that are more consistent with the real HSI. GAN tends to generate spectral curves that are frequently seen in the real     HSI to confuse the discriminator by high-quality samples, so the distribution of generated spectral curves are concentrated. VAE only estimates the main characteristics of the real HSI due to the deficient structure. MPRNet generates high-quality spectral curves, but the distribution is limited to the real HSI, so the generated images lack diversity. UD and UBF combines the physical constraints with reconstruction loss. The spectral distributions show similar patterns with real HSI, but there is a shift in the absolute value. HUD generates high-quality spectral curves that follows the overall shape and distribution of the real HSI, while there are also different distributions in the detail.

\subsection{Quantitative Experiments}

In this section, we will quantitatively analyze the quality of generated images. The evaluation metrics include common IS and FID used in RGB generation tasks, as well as the proposed point fidelity and region diversity. We trained a classification network on each dataset to replace the ImageNet pretrained InceptionV3 used in calculating IS and FID.

\subsubsection{Results on Fidelity and Diversity}

\begin{table*}[htbp]
        \centering
        \caption{Point Fidelity and Block Diversity Comparison.\label{experiment_point_block}}
        \resizebox{1\linewidth}{!}{
        \begin{tabular}{@{}l|cccccc|cccccc|cccccc@{}}
                \bottomrule
                \specialrule{0em}{1.5pt}{1.5pt}
                \midrule
                ~                & \multicolumn{6}{c|}{$F_p$ $\uparrow$} & \multicolumn{6}{c|}{$D_b$ $\downarrow$} & \multicolumn{6}{c}{$D_b/F_p$ $\downarrow$}                                                                                                                                                \\
                \midrule
                \textbf{Dataset} & \textbf{GAN} & \textbf{VAE} & \textbf{MPRNet} & \textbf{UD} & \textbf{UBF} & \textbf{HUD} & \textbf{GAN} & \textbf{VAE} & \textbf{MPRNet} & \textbf{UD} & \textbf{UBF} & \textbf{HUD} & \textbf{GAN} & \textbf{VAE} & \textbf{MPRNet} & \textbf{UD} & \textbf{UBF} & \textbf{HUD} \\
                \midrule
                Indianpines      & 0.996 & 0.914 & 0.996 & 0.558 & 0.559 & 0.999 & 0.981 & 0.914 & 0.995 & 0.553 & 0.553 & 0.978 & 0.984 & 1.000 & 0.999 & 0.991 & 0.989 & \textbf{0.979} \\
                KSC              & 0.998 & 0.978 & 0.983 & 0.836 & 0.831 & 0.999 & 0.988 & 0.978 & 0.983 & 0.833 & 0.830 & 0.990 & \textbf{0.989} & 1.000 & 1.000 & 0.997 & 0.999 & 0.991 \\
                Pavia            & 0.996 & 0.941 & 0.999 & 0.867 & 0.836 & 0.999 & 0.949 & 0.936 & 0.942 & 0.818 & 0.795 & 0.912 & 0.953 & 0.995 & 0.943 & 0.943 & 0.951 & \textbf{0.913} \\
                PaviaU           & 0.996 & 0.941 & 0.997 & 0.875 & 0.866 & 0.999 & 0.948 & 0.934 & 0.984 & 0.823 & 0.820 & 0.919 & 0.952 & 0.992 & 0.987 & 0.941 & 0.947 & \textbf{0.920} \\
                Salinas          & 0.995 & 0.998 & 0.998 & 0.553 & 0.619 & 0.999 & 0.984 & 0.982 & 0.988 & 0.545 & 0.610 & 0.981 & 0.989 & 0.984 & 0.990 & 0.986 & 0.985 & \textbf{0.982} \\
                Avg              & 0.994 & 0.954 & 0.995 & 0.537 & 0.742 & 0.999 & 0.968 & 0.949 & 0.978 & 0.714 & 0.722 & 0.958 & 0.969 & 0.994 & 0.984 & 0.972 & 0.974 & \textbf{0.959} \\
                \hline
                \specialrule{0em}{1.5pt}{1.5pt}
                \bottomrule
        \end{tabular}
        }
\end{table*}

\begin{table}[htbp]
        \centering
        \caption{Inception Score Comparison.\label{experiment_IS}}
        \resizebox{1\linewidth}{!}{
        \begin{tabular}{@{}clcccccc@{}}
                \bottomrule
                \specialrule{0em}{1.5pt}{1.5pt}
                \midrule
                \textbf{Metric}                 & \textbf{Dataset} & \textbf{GAN}  & \textbf{VAE} & \textbf{MPRNet} & \textbf{UD} & \textbf{UBF} & \textbf{HUD}  \\
                \midrule
                \multirow{6}{*}{IS $\uparrow$ } & Indianpines      & \textbf{1.08} & 1.02         & 1.04            & 1.00              & 1.00         & {1.07}        \\
                ~                               & KSC              & 1.00          & 1.00         & 1.00            & 1.00              & 1.00         & 1.00          \\
                ~                               & Pavia            & 1.01          & 1.01         & \textbf{1.04}   & 1.00              & 1.00         & 1.01          \\
                ~                               & PaviaU           & 1.03          & 1.03         & {1.04}          & 1.00              & 1.00         & \textbf{1.05} \\
                ~                               & Salinas          & \textbf{1.05} & {1.04}       & 1.00            & 1.00              & 1.01         & \textbf{1.05} \\
                ~                               & Avg              & {1.03}        & 1.02         & {1.03}          & 1.00              & 1.00         & \textbf{1.04} \\
                \hline
                \specialrule{0em}{1.5pt}{1.5pt}
                \bottomrule
        \end{tabular}
        }
\end{table}

\begin{table}[htbp]
        \centering
        \caption{Fréchet Inception Distance Comparison.\label{experiment_FID}}
        \resizebox{1\linewidth}{!}{
        \begin{tabular}{@{}clcccccc@{}}
                \bottomrule
                \specialrule{0em}{1.5pt}{1.5pt}
                \midrule
                \textbf{Metric}                    & \textbf{Dataset} & \textbf{GAN}   & \textbf{VAE} & \textbf{MPRNet} & \textbf{UD} & \textbf{UBF} & \textbf{HUD}  \\
                \midrule
                \multirow{6}{*}{FID $\downarrow$ } & Indianpines      & 8.03           & 240.91       & 20.98           & 124.41            & 171.62       & \textbf{4.71} \\
                ~                                  & KSC              & 10.87          & {5.45}       & \textbf{4.76}   & 51.99             & 53.68        & 6.69          \\
                ~                                  & Pavia            & 4.81           & 13.82        & \textbf{2.37}   & 23.37             & 19.01        & {4.79}        \\
                ~                                  & PaviaU           & 4.92           & 9.13         & {3.13}          & 14.46             & 11.32        & \textbf{2.81} \\
                ~                                  & Salinas          & \textbf{15.03} & {21.33}      & 29.09           & 76.63             & 91.64        & 23.12         \\
                ~                                  & Avg              & {8.73}         & 58.13        & 12.07           & 58.17             & 69.45        & \textbf{8.42} \\
                \hline
                \specialrule{0em}{1.5pt}{1.5pt}
                \bottomrule
        \end{tabular}
        }
\end{table}

The quantitative results are shown in \cref{experiment_IS,experiment_FID,experiment_point_block}. According to \cref{experiment_IS}, all models achieve similar results while HUD performs slightly better than the other models. According to \cref{experiment_FID}, the results are unstable across different datasets and models. IS and FID are originally proposed for RGB images, which requires a large-scale pretrained InceptionV3. However, the hyperspectral images have significant differences in distribution, so the classification networks should be trained on each dataset. As a result, the metrics are unstable and questionable. Regardless, HUD still achieves the top results in the IS and FID comparison.

As shown in \cref{experiment_point_block}, HUD achieves the highest point fidelity, indicating that its generated images better preserve the spectral content of real images. In contrast, although UD and UBF introduce the physical constraints into data-driven autoencoders, they cannot generate high quality spectral pixels. This highlights the effectiveness of directly applying unmixing algorithm in HSI generation. MPRNet, a spectral super-resolution model, ranks second in point fidelity. UD, UBF and VAE generates the most diverse images, but their poor point fidelity result in overall low-quality generations. HUD achieves the better block diversity compared with GAN and MPRNet, demonstrating richer spatial distributions. MPRNet ranks lowest in block diversity, suggesting its generated images are overly constrained by the original spatial patterns. Combining both point and block diversity, HUD delivers the best overall performance, followed by GAN. This indicates that HUD not only generates high-quality images but also maintains a diverse distribution.

\section{Conclusion and Discussion\label{conclusion}}

We presented an unconditional hyperspectral image (HSI) generation framework that jointly advances fidelity and diversity while addressing two core challenges of HSI synthesis: high spectral dimensionality and physics-driven constraints. Our approach couples a hyperspectral unmixing autoencoder with a diffusion-based generator. The unmixing module projects HSIs into a low-dimensional abundance space, enabling efficient modeling, and the diffusion module operates in a constraint-relaxed domain with a differentiable projection that guarantees non-negativity and sum-to-one upon reconstruction. We further introduced HSI-specific evaluation metrics that better reflect the spectral-spatial characteristics of unconditional generation. Empirically, our model surpasses existing approaches across both proposed and conventional metrics, indicating its effectiveness for realistic and diverse HSI synthesis.

HUD leverages unmixing to impose a physically meaningful latent structure for diffusion. Nevertheless, the data-sparse and distribution-shifted nature of HSI collections poses practical limitations. To mitigate overfitting, we refrain from extensively fine-tuning the unmixing-initialized autoencoder, which constrains adaptability across scenes. Moreover, under current data and computational limit, we train models on a per-image (or per-scene) basis rather than a single, universal generator. The full-resolution, large-footprint HSI synthesis remains computationally prohibitive despite dimensionality reductions. Future work will pursue: (1) A unified generative paradigm capable of accommodating varying numbers of spectral bands and sensor characteristics. (2) Enhanced spatial modeling, including principled spatial compression and multiscale tiling strategies to approach full-resolution synthesis. (3) Data-efficient training protocols (e.g., cross-scene pretraining, self-supervision, and physics-informed regularization) to broaden generalization while preserving physical consistency.

\bibliography{HUD}
\bibliographystyle{IEEEtran}

\begin{IEEEbiographynophoto}{Shiyu Shen} received the B.S. degree from School of Mathematic Science, Nankai University, Tianjin, China, in 2021. He is currently working toward the Ph.D. degree in School of Statistics and Data Science, Nankai University. His research interests include machine learning, representation learning and uncertainty estimation.
\end{IEEEbiographynophoto}

\begin{IEEEbiographynophoto}{Bin Pan} received the B.S. and Ph.D. degrees from the School of Astronautics, Beihang University, Beijing, China, in 2013 and 2019, respectively. Since 2019, he has been an Associate Professor with School of Statistics and Data Science, Nankai University. His research interests include machine learning, remote sensing image processing and multi-objective optimization.
\end{IEEEbiographynophoto}

\begin{IEEEbiographynophoto}{Zifeng Yang} received the B.S. degree from Nankai University, Tianjin, China, in 2023. He is currently working toward the M.S. degree at the School of Statistics and Data Science, Nankai University. His research interests include machine learning, image generation, and diffusion models.
\end{IEEEbiographynophoto}

\begin{IEEEbiographynophoto}{Zhenwei Shi} received the Ph.D.degree in mathematics from the Dalian University of Technology, Dalian, China, in 2005. He was a Post-Doctoral Researcher with the Department of Automation, Tsinghua University, Beijing, China, from 2005 to 2007. He was a Visiting Scholar with the Department of Electrical Engineering and Computer Science, Northwestern University, Evanston, IL, USA, from 2013 to 2014. He is currently a Professor and the Dean of the Image Processing Center, School of Astronautics, Beihang University, Beijing. He has authored or coauthored over 200 scientific articles in refereed journals and proceedings, including the IEEE Transactions on Pattern Analysis and Machine Intelligence, the IEEE Transactions ON Image Processing, the IEEE Transanctions on Geoscience and Remote Sensing, the IEEE Geoscience and Remote Sensing Letters, the IEEE Conference on Computer Vision and Pattern Recognition (CVPR), and the IEEE International Conference on Computer Vision (ICCV). His research interests include remote sensing image processing and analysis, computer vision, pattern recognition, and machine learning. Dr. Shi serves as an Editor for the IEEE Transactions on Geoscience and Remote Sensing, the Pattern Recognition, the ISPRS Journal of Photogrammetry and Remote Sensing, and the Infrared Physics and Technology. His personal website is http://levir.buaa.edu.cn/.
\end{IEEEbiographynophoto}


\end{document}